\documentclass[11pt]{article}

\PassOptionsToPackage{table}{xcolor}
\usepackage[preprint]{acl}

\usepackage{times}
\usepackage{latexsym}
\usepackage[T1]{fontenc}
\usepackage[utf8]{inputenc}
\usepackage{microtype}
\usepackage{inconsolata}
\usepackage{graphicx}
\graphicspath{{./}{figures/}}
\usepackage{booktabs}
\usepackage{amsmath}
\usepackage{algorithm}
\usepackage{algpseudocode}
\usepackage[skins,breakable]{tcolorbox}
\usepackage{placeins}

\definecolor{softblue}{HTML}{EAF3FC}
\definecolor{midblue}{HTML}{D3E5F6}
\definecolor{deepblue}{HTML}{1F5F99}
\definecolor{softgreen}{HTML}{E8F6EE}
\definecolor{midgreen}{HTML}{CFEBDA}
\definecolor{deepgreen}{HTML}{1F7A4A}
\definecolor{softred}{HTML}{FBE3E1}
\newtcolorbox{findingbox}{enhanced,breakable,colback=softgreen,colframe=deepgreen,boxrule=0.4pt,
  leftrule=2.2pt,arc=1.5pt,left=4pt,right=4pt,top=2pt,bottom=2pt,fontupper=\small}

\newtcolorbox{promptbox}[2][]{enhanced,breakable,colback=#2!40,colframe=#2!70!black,boxrule=0.5pt,
  arc=2pt,left=5pt,right=5pt,top=3pt,bottom=3pt,fontupper=\footnotesize,#1}

\title{Beyond Aggregate Accuracy: Item-Level Transitions in Prompt Optimization}

\author{
  Chinmoy Mitra\textsuperscript{1} \quad Tarek Mahmud\textsuperscript{2} \\
  \vspace{0.5em}
  \small \textsuperscript{1}Department of Computer Science \& Engineering, Rajshahi University of Engineering \& Technology, Rajshahi, Bangladesh \\
  \small \textsuperscript{2}Texas A\&M University--Kingsville, Kingsville, Texas, USA
}

\begin{document}
\maketitle
\begin{abstract}
Automatic prompt optimization (APO) uses feedback from model responses to improve prompts automatically. However, its effectiveness is usually measured through overall scores, which do not reveal which questions become correct, which previously correct answers become wrong, or whether improvements extend to unseen questions. We introduce Reflective Feedback-Driven Prompt Optimization (RFDPO), a tool that tracks recoveries (incorrect to correct), regressions (correct to incorrect), and churn (all changed outcomes). RFDPO uses this feedback to guide subsequent revisions and records prompt changes throughout optimization. We evaluate RFDPO across five benchmarks covering law, knowledge, instruction following, mathematics, and privacy, using GPT-5 as the optimizer and GPT-4o-mini and GPT-4.1 Mini as solvers. Our results show that overall accuracy can hide substantial changes: in five of nine configurations, churn was at least five times the absolute net change, and a single-item net change concealed up to 17 changed outcomes. Average validation gains were 7.9 percentage points, compared with 5.2 points on held-out tests. An exploratory ablation showed that item-level feedback reduced churn but did not improve test accuracy. These findings motivate reporting recovery, regression, churn, and performance on unseen data alongside overall accuracy.
\end{abstract}

\section{Introduction}
\label{sec:intro}

\begin{figure*}[!t]
\centering
\includegraphics[width=\textwidth]{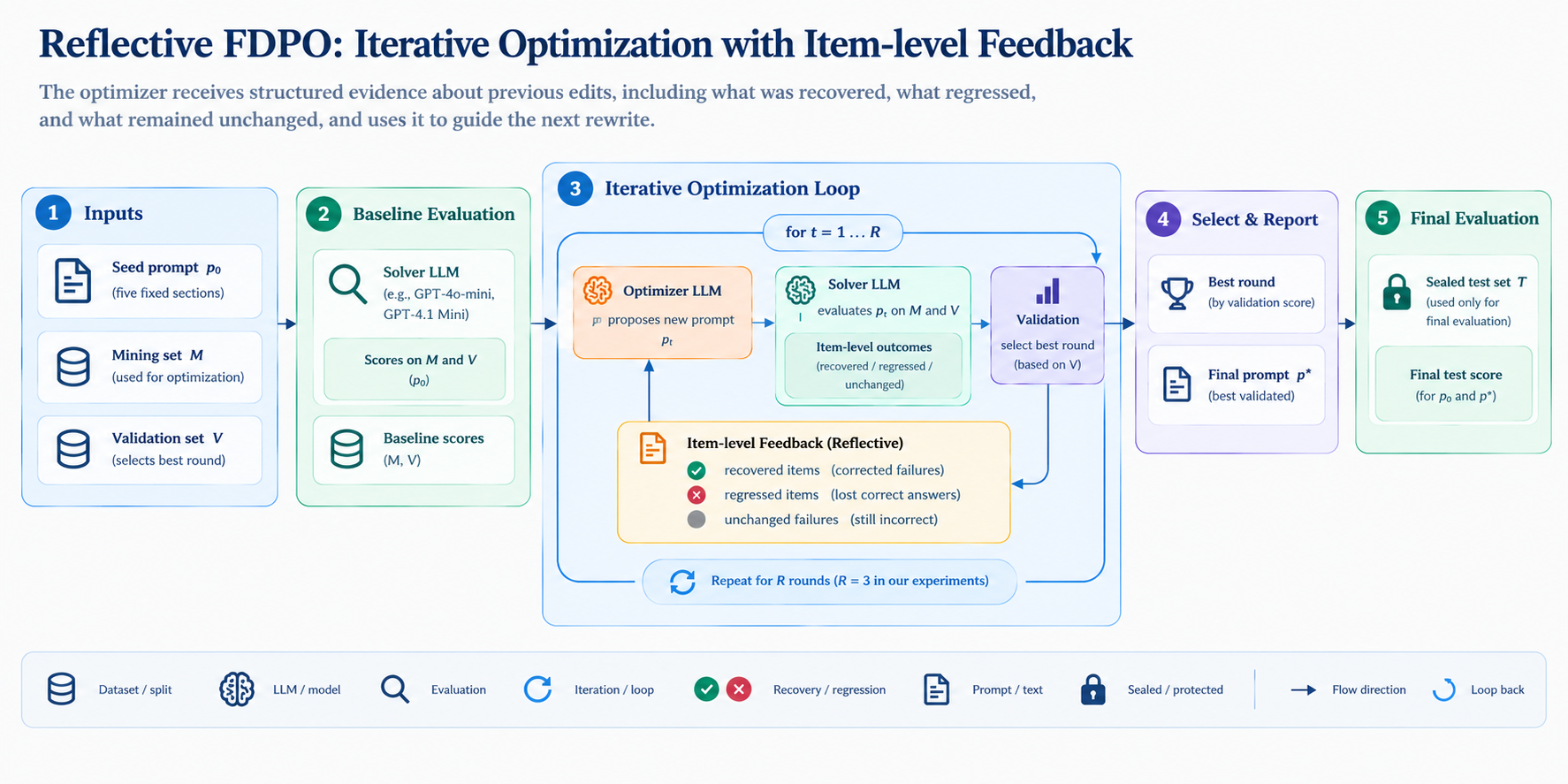}
\caption{The RFDPO loop. The solver evaluates prompts on mining set $M$ and validation set $V$. From the second round,
the optimizer receives the item-level effect of its previous edit before proposing a new prompt. Every changed
candidate is committed, and RFDPO returns the candidate with the highest validation accuracy, breaking ties by mining
accuracy. Test set $T$ is excluded from feedback and selection and evaluated only for the seed and the returned prompt.}
\label{fig:arch}
\end{figure*}

Large language models (LLMs) are increasingly used for document classification, information extraction, question answering, and instruction-following tasks. Their behavior is commonly controlled through natural-language prompts that specify task instructions, relevant context, constraints, and expected output formats. However, even minor changes in prompt wording or structure can substantially affect model performance~\citep{sclar2024quantifying}. Developing effective prompts therefore often requires repeated revisions and evaluations. Automatic prompt optimization (APO) seeks to automate this process by iteratively refining prompts using feedback from model outputs~\citep{zhou2023ape,yang2024opro,pryzant2023protegi}. In LLM-based APO, an optimizer proposes prompt revisions based on evaluation feedback, while a solver executes the task using the revised instructions.

Despite advances in APO, evaluating optimization effectiveness remains challenging. Existing studies primarily report aggregate performance before and after optimization, such as accuracy improvements on benchmark datasets. Although these metrics indicate whether a revised prompt performs better overall, they do not reveal which previously incorrect examples were fixed or which correct examples became incorrect. This distinction is important for practical LLM applications, where prompt revisions may introduce unexpected failures even when overall performance improves. For example, \citet{ma2024regression} found that 10.9\% of individual predictions regressed across hosted LLM API updates for toxicity detection, and previously correct predictions broke in 87.9\% of cases where aggregate accuracy improved. Developers have also reported difficulties debugging inconsistent LLM behavior across prompt variations~\citep{errica2025wrong}. These observations highlight the importance of examining behavioral changes beyond aggregate scores.

Consider two prompt revisions that each improve accuracy by two examples. The first may correct two previously incorrect examples without introducing new failures, while the second may correct ten examples but break eight previously correct ones. Although both revisions achieve the same net improvement, their effects on individual examples differ substantially. Aggregate accuracy alone cannot distinguish these outcomes, making it difficult to assess the trade-offs introduced by optimization or identify which changes should inform subsequent revisions.

Another concern is whether improvements observed during optimization generalize to unseen examples. APO methods commonly select prompts using validation performance, and some also expose validation examples as optimization feedback. Consequently, improvements on these tuning examples may not translate into comparable improvements on held-out data~\citep{recht2019imagenet}. Understanding prompt optimization therefore requires examining both item-level behavioral changes and the relationship between tuning and test performance.

To address these concerns, we introduce \emph{Reflective Feedback-Driven Prompt Optimization} (RFDPO), an auditable prompt optimization tool that explicitly tracks the effects of individual prompt revisions (Figure~\ref{fig:arch}). RFDPO identifies \emph{recoveries} (incorrect-to-correct transitions), \emph{regressions} (correct-to-incorrect transitions), and \emph{churn} (the total number of changed outcomes). Unlike approaches that primarily use aggregate scores, current errors, or execution traces to guide revisions, RFDPO returns the observed consequences of the preceding edit to the optimizer. Beginning with the second round, this feedback identifies recovered and regressed examples, their associated outputs, and the prompt sections modified by the previous revision. RFDPO also records prompt versions, evaluation outcomes, and resource consumption to support analysis of optimization trajectories.

We evaluate RFDPO on five benchmarks---LegalBench-Hearsay, MMLU, IFBench, AIME, and PUPA---using GPT-4o-mini and GPT-4.1 Mini as solver models. Our results show that aggregate accuracy can conceal substantial behavioral changes: in five of nine configurations, churn was at least five times the absolute net change, and a single-item net change concealed up to 17 changed outcomes. Across edited-prompt returns, average held-out test improvements were approximately 61\% of average validation improvements. We further examine how prompt revisions evolve across optimization rounds. An exploratory ablation on LegalBench-Hearsay shows that providing item-level feedback is associated with approximately 40\% lower mining-set churn at similar net gains, although it does not improve test accuracy.

Our main contributions are:

\begin{itemize}
\setlength{\itemsep}{2pt}
\setlength{\parskip}{0pt}
\setlength{\topsep}{2pt}

\item \textbf{An auditable prompt optimization tool.}
We develop RFDPO, which incorporates item-level recovery and regression feedback into iterative prompt revision and records prompt versions, evaluation outcomes, and resource consumption (\S\ref{sec:method}).

\item \textbf{An empirical characterization of prompt revisions.}
We analyze recovery, regression, and churn across five benchmarks, demonstrating how aggregate accuracy conceals changes in individual outcomes and how validation improvements differ from held-out test performance (\S\ref{sec:results}).

\item \textbf{An analysis of feedback and editing behavior.}
We examine prompt modifications across optimization rounds and conduct an exploratory ablation to investigate how item-level feedback affects outcome churn. Based on these findings, we discuss implications for evaluating and reporting APO behavior (\S\ref{sec:edits}, \S\ref{sec:discussion}).

\end{itemize}

Overall, our findings motivate incorporating recovery, regression, churn, and validation-to-test transfer into the evaluation of automatic prompt optimization, alongside conventional aggregate performance metrics.

\section{Related Work}
\label{sec:related}

\paragraph{Automatic prompt optimization.}
Automatic prompt optimization (APO) has been explored through discrete token search~\citep{shin2020autoprompt}, instruction generation and selection~\citep{zhou2023ape}, LLM-based optimization~\citep{yang2024opro}, textual gradients~\citep{pryzant2023protegi,yuksekgonul2025textgrad}, and evolutionary approaches~\citep{guo2024evoprompt,fernando2024promptbreeder}. Other methods optimize multi-stage pipelines or structured prompts~\citep{opsahlong2024miprov2,schnabel2024sammo,khattab2024dspy}. Among these, MPO~\citep{sharma2026mpo} uses the same five-section prompt structure as RFDPO and applies section-local textual gradients. More closely related to our feedback mechanism, GEPA~\citep{agrawal2026gepa} uses reflective prompt mutation and Pareto-based selection, while Trace2Policy~\citep{zha2026trace2policy} clusters errors and applies a regression gate when refining decision rules. Self-feedback methods also guide revisions through verbal critiques~\citep{madaan2023selfrefine,shinn2023reflexion}. While existing methods use scored histories, execution traces, or regression checks to guide optimization, RFDPO explicitly returns the observed item-level consequences of the preceding edit to the optimizer, identifying recovered and regressed examples and the prompt sections modified. Unlike regression-gated approaches, RFDPO retains changed candidates even when validation performance decreases, allowing subsequent revisions to build on earlier edits while recording their effects.

\paragraph{Behavioral evaluation and regression tracking.}
Aggregate accuracy can conceal important changes in model behavior. CheckList~\citep{ribeiro2020checklist} demonstrates the limitations of aggregate evaluation, while negative flips have motivated regression-free model updating~\citep{yan2021positive}. Paired outcome comparisons also have a long history in statistical evaluation~\citep{mcnemar1947sampling}. For LLM applications, \citet{ma2024regression} highlight how previously correct behavior can break following API updates. Related studies show that small prompt changes can alter model predictions~\citep{salinas2024butterfly}, motivating sensitivity and consistency measures~\citep{errica2025wrong}. However, some apparent prompt sensitivity may result from evaluation artifacts~\citep{hua2025flaw}. Unlike evaluations of prompt perturbations or model updates, our work tracks labeled recoveries and regressions across successive prompt optimization revisions using benchmark-specific correctness criteria. We additionally measure churn to distinguish the extent of behavioral change from the net accuracy difference. This distinction is particularly relevant when several recovered examples are offset by regressions, leaving aggregate performance nearly unchanged.

\paragraph{Evaluation protocols and generalization.}
Prompt optimization results depend on evaluation settings, including data splits, solver models, baseline prompts, and feedback availability. Consequently, published APO scores are not directly comparable across different protocols (Appendix~\ref{app:prior}, Table~\ref{tab:prior}). Moreover, improvements on tuning data may not generalize to unseen examples~\citep{recht2019imagenet}, and benchmark contamination presents additional evaluation concerns~\citep{oren2023proving}. Test-set exposure is particularly relevant: Trace2Policy reports a refinement round informed by held-out errors~\citep{zha2026trace2policy}. In our diagnostic experiment, providing test examples as feedback yielded an accuracy of 95.3\%, compared with 73.4\% when they were withheld (Appendix~\ref{app:oracle}). This diagnostic illustrates the importance of distinguishing optimization feedback from held-out evaluation. RFDPO excludes test examples from feedback and prompt selection within each optimization run. Because validation examples also contribute feedback, we treat validation performance as a tuning measure and separately report performance changes on test examples to examine generalization.

\section{RFDPO}
\label{sec:method}

We introduce Reflective Feedback-Driven Prompt Optimization (RFDPO) that uses item-level evaluation outcomes to guide prompt revisions. RFDPO extends conventional feedback-driven optimization by explicitly identifying examples that become correct or incorrect after each revision and returning these changes to the optimizer. Figure~\ref{fig:arch} illustrates the overall workflow.

\subsection{Prompt Representation and Evaluation}

RFDPO uses a solver LLM to execute tasks and an optimizer LLM to revise prompts, without modifying the solver's parameters. Following MPO~\citep{sharma2026mpo}, each prompt contains five sections: System Role, Context, Task Details, Constraints, and Output Format. These sections remain fixed throughout optimization, allowing RFDPO to track which parts of a prompt are modified.

The tool operates on three disjoint sets: mining ($M$), validation ($V$), and test ($T$). Mining examples provide feedback for prompt revisions, while validation examples are used for candidate selection and, from the second round, also contribute feedback. The test set is excluded from optimization and evaluated only with the initial and returned prompts.

\subsection{Item-Level Transition Tracking}
\label{sec:metric}

RFDPO measures the effect of each revision by comparing example-level correctness before and after an edit. Let $c_i(p)\in\{0,1\}$ denote the observed correctness of example $i$ under prompt $p$. Given two prompts $p$ and $q$ evaluated on the same example set $D$, we define the recovered ($R$) and regressed ($G$) examples as
\vspace{-8px}
\begin{equation}
R=\{i\in D:c_i(p)=0,\ c_i(q)=1\},
\end{equation}
\begin{equation}
G=\{i\in D:c_i(p)=1,\ c_i(q)=0\}.
\end{equation}

We further define the net change $N$ and churn $C$ as
\vspace{-8px}
\begin{equation}
N=|R|-|G|,\qquad C=|R|+|G|.
\end{equation}

\begin{figure}[!t]
\centering
\includegraphics[width=\columnwidth]{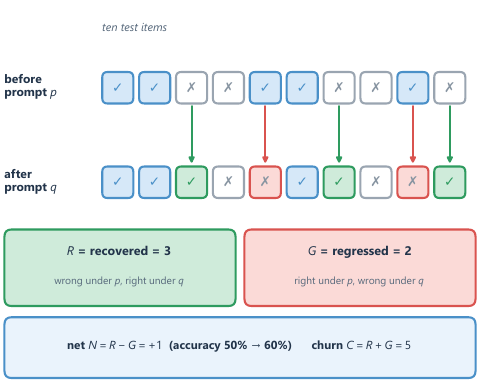}
\caption{Example of item-level transitions. Three examples are recovered and two regress, producing a net gain of one correct example and a churn of five.}
\label{fig:metric}
\end{figure}

The net change corresponds to $|D|$ times the accuracy difference, whereas churn captures the total number of examples whose correctness changes. Thus, revisions with the same accuracy change may exhibit substantially different recovery and regression patterns, as illustrated in Figure~\ref{fig:metric}.

Transitions are measured between consecutive revisions on $M$ and $V$, and between the initial and returned prompts on $T$, using matched evaluable example IDs.

\subsection{Feedback-Driven Prompt Revision}
\label{sec:optimization}

Algorithm~\ref{alg:reflect} presents the RFDPO optimization procedure. The solver first evaluates the initial prompt, and the optimizer generates a revision using the observed outcomes. From the second round, the optimizer additionally receives the recoveries and regressions caused by the preceding revision on $M$ and $V$, together with the affected questions, solver responses, gold answers, previous text of modified sections, and validation accuracy changes.

Each valid, changed candidate is evaluated and committed as the starting point for the next round, even if its validation accuracy decreases. This allows subsequent revisions to build on earlier edits without a regression-based acceptance gate.

Under best-validation selection, RFDPO returns the committed candidate with the highest validation accuracy, breaking ties by mining accuracy and candidate order. Once a candidate is committed, the seed is not reconsidered; if no candidate is committed, the seed is retained.

RFDPO also records prompt versions, modified sections, item-level outcomes, and model-call token usage and costs. These records support analysis of optimization trajectories, behavioral changes, and resource consumption.

\begin{algorithm}[!t]
\caption{RFDPO optimization with edit-effect feedback}
\label{alg:reflect}
\small
\begin{algorithmic}[1]
\Require seed $p_0$; disjoint $M,V,T$; $\textsc{Opt}$; $\textsc{Solve}$; round budget $B$
\State $s_0,v_0 \gets \textsc{Solve}(p_0,M),\ \textsc{Solve}(p_0,V)$
\State $\tau_0 \gets \textsc{Solve}(p_0,T)$ \Comment{baseline; measured only}
\State $\mathcal{H}\gets\emptyset$; $(p,s,v)\gets(p_0,s_0,v_0)$; $\rho\gets\emptyset$
\For{$t=1$ \textbf{to} $B$}
  \If{skip threshold reached, or no mining failures} \State \textbf{break} \EndIf
  \State $q\gets\textsc{Opt}(p,s,v,\rho)$ \Comment{mining items and feedback}
  \If{$q=p$} \State \textbf{continue} \EndIf
  \State $s',v'\gets\textsc{Solve}(q,M),\ \textsc{Solve}(q,V)$
  \State $\rho\gets((R,G)(s\!\to\!s'),(R,G)(v\!\to\!v'))$
  \State append $(q,s',v')$ to $\mathcal{H}$ \Comment{no regression gate}
  \State $(p,s,v)\gets(q,s',v')$
\EndFor
\State $p^*\gets p_0$ if $\mathcal{H}=\emptyset$; otherwise select from $\mathcal{H}$
\Statex \hspace{\algorithmicindent} by validation accuracy, then mining accuracy
\State $\tau^* \gets \textsc{Solve}(p^*,T)$ \Comment{measured only}
\State \Return $p^*$, and $(R,G)(\tau_0\!\to\!\tau^*;T)$
\end{algorithmic}
\end{algorithm}

\section{Experimental Setup}
\label{sec:setup}

We evaluate 5 benchmarks using RFDPO and 2 solver models, covering different task types, answer formats, and scoring criteria. Table~\ref{tab:setup} summarizes the benchmark configurations.

\subsection{Benchmarks}

We select five benchmarks covering diverse tasks, including legal reasoning, general knowledge, instruction following, mathematical reasoning, and privacy preservation.

\textbf{LegalBench-Hearsay}~\citep{guha2023legalbench} evaluates binary classification of statements under U.S. hearsay law.

\textbf{MMLU}~\citep{hendrycks2021mmlu} consists of four-way multiple-choice questions. We select six subjects: college mathematics, computer security, econometrics, high school biology, philosophy, and professional law, optimizing each independently and reporting macro-averaged accuracy.

\textbf{IFBench}~\citep{pyatkin2025ifbench} evaluates compliance with verifiable output constraints. An example is correct only when all applicable constraints are satisfied, using a programmatic verifier. ~\citep{zhou2023ifeval}.

\textbf{AIME}~\citep{maa2025aime} evaluates mathematical reasoning with integer answers between 0 and 999. We use problems from 2022--2024 for optimization and the 30 problems from 2025 for testing.

\textbf{PUPA}~\citep{li2025papillon} evaluates privacy-conscious delegation to an external LLM. We optimize only the query-redaction prompt. Correctness requires a composite score $(\mathrm{quality}+1-\mathrm{leakage})/2$ of at least $0.7$.

\begin{table*}[!t]
\centering
\small
\resizebox{\linewidth}{!}{%
\begin{tabular}{lllcc}
\toprule
\rowcolor{midblue}
Benchmark & Task & Reported metric & Train $\rightarrow$ $M$/$V$ & Test \\
\midrule
LegalBench-Hearsay & Binary classification & Accuracy & $50\rightarrow 25/25$ & 49 \\
MMLU (6 subjects) & Multiple choice & Accuracy, macro-avg. & $50\rightarrow 25/25$ & 66 each \\
IFBench & Instruction following & Accuracy & $40\rightarrow 26/14$ & 42 \\
AIME (2022--24 $\rightarrow$ 2025) & Mathematical reasoning & Accuracy & $90\rightarrow 58/32$ & 30 \\
PUPA & Privacy-conscious delegation & Accuracy (composite $\geq 0.7$) & $60\rightarrow 30/30$ & 40 \\
\bottomrule
\end{tabular}}
\caption{Benchmarks and evaluation splits. MMLU splits are per subject. Test sizes precede provider-block exclusions.}
\label{tab:setup}
\end{table*}

\subsection{Models and Experimental Protocol}

We use GPT-4o-mini~\citep{openai2024gpt4o} and GPT-4.1 Mini~\citep{openai2025gpt41} as solvers, both with temperature $0$, and GPT-5~\citep{openai2025gpt5} as the optimizer with temperature $0.7$. PUPA additionally uses a fixed GPT-4.1 external model and GPT-5 as its quality judge.

All experiments use seed $0$ and at most three optimization rounds. Optimization is skipped when initial mining accuracy reaches $95\%$, except for PUPA. 

The evaluation includes 19 baseline/final records: seven non-MMLU configurations and twelve independently optimized MMLU subjects. Both endpoints of each matched pair are evaluated within the same run. Two earlier AIME runs, which used a 2,048-token output limit, are reported separately in Appendix~\ref{app:aime_extra} and excluded from the analyses.

We report test accuracy, validation accuracy changes, recovery, regression, and churn. We also examine revision-level transitions and recorded token consumption. Detailed run configurations, scoring procedures, and optimizer instructions are provided in Appendices~\ref{app:config}, \ref{app:eval}, and \ref{app:prompts}.

\section{Results}
\label{sec:results}

\subsection{Item-Level Performance Changes}
\label{sec:itemresults}

\begin{figure*}[!t]
\centering
\includegraphics[width=0.94\textwidth]{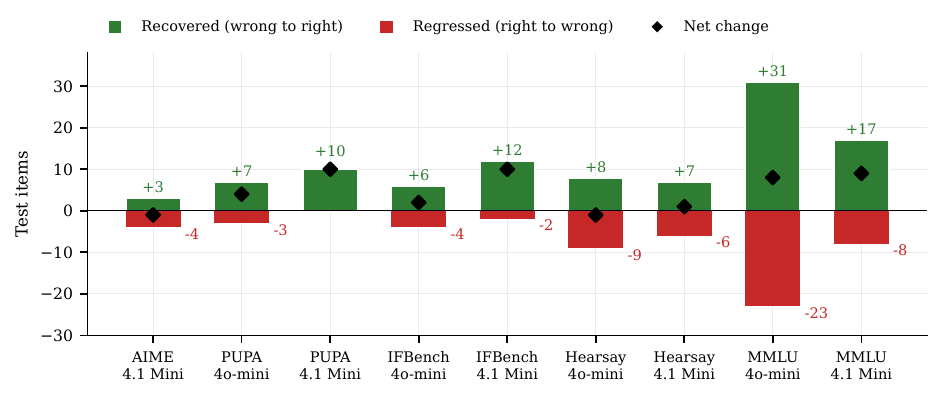}
\caption{Recovered and regressed test examples across benchmarks and solvers. Diamonds indicate net changes. MMLU transitions are aggregated across six subjects.}
\label{fig:rr}
\end{figure*}

We first examine how prompt optimization affects individual test examples. Table~\ref{tab:main} reports test accuracy before and after optimization, validation accuracy changes, and the number of recovered and regressed examples. Figure~\ref{fig:rr} visualizes these transitions, while Appendix~\ref{app:gap} provides the individual accuracy records.

\begin{table*}[!t]
\centering
\small
\setlength{\tabcolsep}{4pt}
\begin{tabular}{llrrrrrrr}
\toprule
\rowcolor{midblue}
Benchmark & Solver & \multicolumn{3}{c}{Test accuracy (\%)} &
\shortstack{$\Delta V$\\(pp)} &
\multicolumn{3}{c}{Test transitions} \\
\cmidrule(lr){3-5}\cmidrule(lr){7-9}
& & Seed & Final & $\Delta T$ & & $|R|$ & $|G|$ & Churn (\%) \\
\midrule
AIME & GPT-4.1 Mini & 53.3 & 50.0 & $-3.3$ & $+6.2$ & 3 & 4 & 7 (23.3) \\
\rowcolor{softgreen}
PUPA & GPT-4o-mini & 55.3 & 65.8 & $+10.5$ & $+13.3$ & 7 & 3 & 10 (26.3) \\
\rowcolor{softgreen}
PUPA & GPT-4.1 Mini & 44.7 & 71.1 & $+26.3$ & $+23.3$ & 10 & 0 & 10 (26.3) \\
\rowcolor{softgreen}
IFBench & GPT-4o-mini & 40.5 & 45.2 & $+4.8$ & $+15.8$ & 6 & 4 & 10 (23.8) \\
\rowcolor{softgreen}
IFBench & GPT-4.1 Mini & 42.9 & 66.7 & $+23.8$ & $+14.3$ & 12 & 2 & 14 (33.3) \\
Hearsay & GPT-4o-mini & 71.4 & 69.4 & $-2.0$ & $+12.0$ & 8 & 9 & 17 (34.7) \\
\rowcolor{softgreen}
Hearsay & GPT-4.1 Mini & 71.4 & 73.5 & $+2.0$ & $+20.0$ & 7 & 6 & 13 (26.5) \\
\rowcolor{softgreen}
MMLU (macro) & GPT-4o-mini & 75.1 & 77.2 & $+2.0$ & $+5.5$ & 31 & 23 & 54 (13.7) \\
\rowcolor{softgreen}
MMLU (macro) & GPT-4.1 Mini & 81.0 & 83.3 & $+2.3$ & $-0.7$ & 17 & 8 & 25 (6.4) \\
\bottomrule
\end{tabular}
\caption{Seed-to-returned test accuracy, validation accuracy change, and item-level transitions. Green rows indicate positive test changes. Churn percentages use evaluated test items. MMLU accuracy is macro-averaged across six subjects, while transitions sum 393 items. PUPA (38 items) and professional law (63 items) exclude provider-blocked examples. Earlier AIME runs with other solvers are described in Appendix~\ref{app:aime_extra}.}
\label{tab:main}
\end{table*}

Test accuracy increased in seven of the nine reported configurations, with the largest gain of 26.3 percentage points on PUPA using GPT-4.1 Mini. However, aggregate improvements do not fully represent the changes in individual predictions.

On LegalBench-Hearsay, a net change of only one correct example concealed 13--17 changed outcomes, corresponding to 27--35\% of the test set. Similarly, IFBench with GPT-4.1 Mini recovered 12 examples but regressed two, yielding a net gain of ten. On MMLU with GPT-4o-mini, 54 examples changed correctness despite a net improvement of only eight.

Across five configurations, churn was at least five times the absolute net change. These results show that small accuracy differences can conceal substantial recovery and regression, motivating item-level analysis alongside aggregate performance.

\subsection{Validation-to-Test Transfer}
\label{sec:valtest}

We next examine whether improvements observed during optimization transfer to held-out test examples. Among the 19 records, 17 returned edited prompts. Of these, validation accuracy increased in 12, remained unchanged in three, and decreased in two. Table~\ref{tab:gap} summarizes the corresponding validation and test changes.

\begin{table}[!t]
\centering
\small
\resizebox{\linewidth}{!}{
\begin{tabular}{lrrrrr}
\toprule
\rowcolor{midblue}
Edited-prompt returns & Runs & Val.\ $\uparrow$ &
\shortstack{Val.\ $\uparrow$\\test $\downarrow$} &
$\overline{\Delta V}$ & $\overline{\Delta T}$ \\
\midrule
Non-MMLU & 7 & 7 & 2 & $+15.0$ & $+8.9$ \\
MMLU subjects & 10 & 5 & 1 & $+2.9$ & $+2.6$ \\
\textbf{All matched} & \textbf{17} & \textbf{12} & \textbf{3} &
$\mathbf{+7.9}$ & $\mathbf{+5.2}$ \\
\bottomrule
\end{tabular}}
\caption{Mean validation and test accuracy changes for 17 edited-prompt returns; two additional records retained the seed. Means weight runs equally and are reported in percentage points. The fourth column counts runs with improved validation accuracy but decreased test accuracy.}
\label{tab:gap}
\end{table}

Of the 12 runs with improved validation accuracy, nine also improved test accuracy, while three lost one test example each. These cases occurred on AIME with GPT-4.1 Mini ($+6.2$ validation, $-3.3$ test points), Hearsay with GPT-4o-mini ($+12.0$, $-2.0$), and MMLU computer security with GPT-4o-mini ($+8.0$, $-1.5$).

Across the 17 edited-prompt returns, validation accuracy improved by 7.9 percentage points on average, compared with 5.2 points on test data. The test gain was therefore approximately 66\% of the average validation gain. The difference was smaller for MMLU ($+2.9$ validation versus $+2.6$ test) than for the other benchmarks ($+15.0$ versus $+8.9$).
Since validation examples also provide optimization feedback, these results describe the transfer of tuning gains to held-out examples rather than improvements on an independent validation set.

\paragraph{Verbatim overlap.}
Because the optimizer sees every failure and success on its working set, it could inflate tuning accuracy by pasting those items, or paraphrases of them, into the prompt. Its instructions therefore forbid reusing the scenario of any shown item and prefer abstract rules (Appendix~\ref{app:prompts}). We also checked whether returned prompts contained eight-word sequences copied from benchmark examples. Sixteen of the 19 prompts showed no such overlap. The remaining three contained either a generic mathematical phrase or a True/False answer template (Appendix~\ref{app:overlap}). The check detects only verbatim copying; paraphrase is guarded by the instruction alone.

\subsection{Prompt Revision Behavior}
\label{sec:edits}

We analyze how the optimizer modifies prompts across successive rounds. In round 1, where no effect report of a previous edit is available, the optimizer modified an average of 4.3 of the five prompt sections, compared with 1.3 in rounds 2--3, where it receives recovery and regression feedback on its previous edit. Revisions thus became more localized after the initial round.


\paragraph{Effect-report ablation (exploratory).}
To see what the effect report contributes beyond the optimization round, we ran an exploratory ablation on LegalBench-Hearsay using GPT-4o-mini. Both conditions replayed the same round-1 prompt and continued through rounds 2 and 3. They differed in whether the optimizer received the item-level effect report and its accompanying instructions (Appendix~\ref{app:sysprompt}). We performed five runs per condition (Appendix~\ref{app:ablation}).

Without the report, round-2 revisions already modified only 1.6 prompt sections on average, compared with 2.0 when the report was included. Thus, narrower edits also occur without the report.

The report did change the number of changed outcomes. With it, an average of 3.1 mining examples changed correctness per round, compared with 5.2 without it ($p=0.024$, exact permutation test over runs, uncorrected). This average pools rounds 2 and 3, and round 3 starts from different parent prompts in the two conditions. Both conditions achieved similar net gains of 0.3 and 0.2 mining examples per round, respectively.

Test accuracy fell in both conditions, by 5.3 and 5.7 percentage points with and without the report, respectively, so neither condition improved test accuracy. We therefore treat the churn difference as exploratory evidence about what item-level feedback changes in this setting, not as evidence that it improves accuracy.

\subsection{Resource Consumption}
\label{sec:resources}

Finally, we examine the computational resources recorded during optimization. Table~\ref{tab:resources} summarizes optimizer calls, token consumption, and logged inference costs across the matched configurations.

\begin{table*}[!t]
\centering
\small
\setlength{\tabcolsep}{5pt}
\begin{tabular}{llrrrrr}
\toprule
\rowcolor{midblue}
Benchmark & Solver &
\shortstack{Optimizer\\calls} &
\shortstack{Optimizer\\input (K)} &
\shortstack{Optimizer\\output (K)} &
\shortstack{All-role\\tokens (K)} &
\shortstack{Logged cost\\(USD)} \\
\midrule
AIME & GPT-4.1 Mini & 3 & 498.4 & 5.4 & 2,666.6 & 3.81 \\
PUPA & GPT-4o-mini & 3 & 73.8 & 13.7 & 1,651.2 & 4.44 \\
PUPA & GPT-4.1 Mini & 3 & 68.5 & 15.6 & 1,570.4 & 4.80 \\
IFBench & GPT-4o-mini & 3 & 23.4 & 12.0 & 300.9 & 0.21 \\
IFBench & GPT-4.1 Mini & 3 & 20.2 & 10.1 & 240.1 & 0.35 \\
Hearsay & GPT-4o-mini & 3 & 17.1 & 14.9 & 241.3 & 0.22 \\
Hearsay & GPT-4.1 Mini & 3 & 14.1 & 20.5 & 246.6 & 0.37 \\
MMLU (six subjects) & GPT-4o-mini & 18 & 123.0 & 77.1 & 2,014.2 & 1.32 \\
MMLU (six subjects) & GPT-4.1 Mini & 11 & 75.7 & 34.2 & 1,170.8 & 1.10 \\
\bottomrule
\end{tabular}
\caption{Inference consumption across matched configurations. K denotes 1,000 tokens. All-role totals include solver, optimizer, and applicable external-model and judge calls, including seed and final evaluations. Costs use configured prices. MMLU aggregates six subjects; two GPT-4.1 Mini subject runs skipped optimization.}
\label{tab:resources}
\end{table*}

Resource consumption varied considerably across tasks. Three optimizer calls required approximately 498K input tokens for AIME with GPT-4.1 Mini, compared with 14K for Hearsay, a difference of roughly 35 times. PUPA incurred additional costs from external-model responses and quality judgments, while MMLU totals aggregate independent optimization runs across six subjects.

These measurements document the resource requirements of RFDPO across different benchmark configurations.

\section{Discussion}
\label{sec:discussion}

\paragraph{Beyond aggregate accuracy.}
Our findings highlight the limitations of evaluating prompt optimization solely through aggregate accuracy. In five of the nine configurations, churn was at least five times the absolute net change, with a single-item improvement or decline concealing up to 17 changed outcomes. Such changes are important in practice because an optimized prompt may introduce failures on examples that previously worked. Recovery and regression tracking makes these changes visible, allowing developers to distinguish newly corrected examples from newly introduced failures. This aligns with regression testing for LLM applications, where previously correct behavior must be monitored as systems evolve~\citep{ma2024regression}.

\paragraph{Role of item-level feedback.}
The optimizer made more localized revisions in later rounds, but our ablation shows that this behavior cannot be attributed solely to the effect report. Nevertheless, providing item-level feedback reduced average mining-set churn from 5.2 to 3.1 examples per round while maintaining similar net gains. This suggests that explicit recovery and regression feedback may help limit unintended changes without necessarily improving overall accuracy. However, the evidence remains exploratory, as the ablation covers one benchmark and solver with five runs per condition, pools two rounds, and includes additional optimizer instructions with the report. Test accuracy declined in both conditions, so we do not interpret the reduced churn as improved generalization.

\paragraph{Generalization and reporting implications.}
Improvements during optimization do not necessarily transfer to unseen examples. Across edited-prompt returns, average test gains were approximately 61\% of average validation gains, and four runs experienced test declines despite validation improvements. Because validation examples also guide optimization, their performance should be interpreted as tuning results rather than independent evidence of generalization. We therefore recommend reporting recovery, regression, and churn alongside validation and test accuracy, prompt versions, evaluation exclusions, and resource consumption. These measurements can be obtained from existing item-level evaluations without additional solver calls. RFDPO automatically records this information, supporting more transparent analysis of prompt optimization (Appendix~\ref{app:artifacts}).

\section{Conclusion}
Aggregate accuracy captures the overall effect of prompt optimization but obscures which examples improve or regress. We presented RFDPO, an auditable optimizer that tracks item-level recoveries, regressions, and churn and incorporates this feedback into subsequent revisions. Across five benchmarks and two solver models, a single-item net change concealed up to 17 changed outcomes, while churn was at least five times the absolute net change in five of nine configurations. Average held-out test gains were approximately 61\% of validation gains. An exploratory Hearsay ablation showed that item-level feedback reduced churn at similar net gains but did not improve test accuracy. These findings highlight that prompt optimization can introduce failures despite aggregate improvements, motivating the reporting of recovery, regression, churn, and validation-to-test transfer alongside accuracy.

\section*{Limitations}
\label{sec:limitations}

\paragraph{Evaluation scope.}
Our evaluation covers five benchmarks, two closed-weight solvers, and one optimizer, with a single seed and relatively small test sets. We did not evaluate open-weight models or longer optimization trajectories. Differences in selection rules and evaluation configurations also limit cross-model comparisons.

\paragraph{Evaluation variability.}
Observed recovery and regression may partly reflect model output variability rather than prompt changes, as shown by unchanged-prompt re-evaluations. PUPA additionally relies on an LLM judge, while the IFBench verifier supports only a subset of constraint types.

\paragraph{Feedback ablation.}
Our exploratory ablation covers one benchmark and solver, with five runs per condition. It combines two optimization rounds, and the effect report includes additional optimizer instructions. Although feedback reduced mining-set churn, it did not improve test accuracy, and its effect may not generalize.

\paragraph{Generalization and data exposure.}
Validation examples are used for both feedback and selection, potentially inflating tuning gains. Test examples remain excluded from feedback within each run, but small test sets limit generalization claims. Our verbatim-overlap check also cannot detect paraphrased copying or semantic memorization.

\section*{Ethical Considerations}
Data come from public benchmarks, resplit as described in Table~\ref{tab:setup}. PUPA contains personal information from
public user--LLM conversations. Hosted solver and optimizer APIs receive the original queries during evaluation and
feedback; redaction protects only the designated external model. String-matched leakage does not capture paraphrased or
inferred information. Released traces would contain item IDs and outcomes only, without raw inputs, predictions, or
gold text (Appendix~\ref{app:artifacts}).

\FloatBarrier
\bibliography{custom}

\onecolumn
\appendix

\newcommand{\appsec}[2]{\par\vspace{4pt}\noindent\makebox[2em][l]{\bfseries\ref{#1}}{\bfseries #2}\,\dotfill\,{\bfseries\pageref{#1}}\par\vspace{1pt}}
\newcommand{\appsubsec}[2]{\par\noindent\hspace{2em}\makebox[2.4em][l]{\ref{#1}}#2\,\dotfill\,\pageref{#1}\par}

\begin{center}{\large\bfseries Appendix Contents}\end{center}
{\small
\appsec{app:config}{Run Configuration and Scoring}
\appsubsec{app:population}{Analysis Population and Settings}
\appsubsec{app:eval}{Scoring and Evaluable Items}
\appsubsec{app:prior}{Related-Work Protocol Comparison}
\appsec{app:prompts}{Prompt and Feedback Specification}
\appsubsec{app:sysprompt}{Optimizer System Prompt}
\appsubsec{app:insertions}{Task-Specific Insertions}
\appsubsec{app:seed}{Seed Prompt Example}
\appsubsec{app:feedback}{Per-Round Feedback}
\appsec{app:results}{Detailed Results and Trajectories}
\appsubsec{app:gap}{Paired Accuracy and Test Transitions}
\appsubsec{app:rounds}{Round-by-Round Trajectories}
\appsubsec{app:evolution}{Worked Prompt Changes}
\appsubsec{app:ablation}{Effect-Report Ablation}
\appsubsec{app:aime_extra}{Additional AIME Runs}
\appsec{app:noisebugs}{Evaluation Checks and Artifacts}
\appsubsec{app:noise}{Re-evaluation of an Unchanged Prompt}
\appsubsec{app:oracle}{Test-Exposure Diagnostic}
\appsubsec{app:overlap}{Verbatim-Overlap Check}
\appsubsec{app:artifacts}{Artifacts and Reproducibility}
}
\bigskip

\section{Run Configuration and Scoring}
\label{app:config}
\subsection{Analysis Population and Settings}
\label{app:population}
The analysis uses 19 completed records: seven non-MMLU runs and twelve independently optimized MMLU
subjects, taking the latest completed record within each fixed run group.

All recorded solvers use temperature 0.0 and the GPT-5 optimizer uses 0.7, with at most three rounds and one optimizer
call per round. Table~\ref{tab:config} gives the configured solver output limits and splits for both solvers.
The response ledgers identify GPT-4o-mini-2024-07-18, GPT-4.1-mini-2025-04-14, and GPT-5-2025-08-07; some entries retain
model aliases rather than dated identifiers. PUPA additionally uses GPT-4.1 as its fixed external model and GPT-5 as judge.

\begin{table}[htbp]
\centering
\small
\setlength{\tabcolsep}{5pt}
\begin{tabular}{llccr}
\toprule
\rowcolor{midblue}Benchmark & Solver & Assigned $M/V$ & Val.\ fraction & Solver limit \\
\midrule
AIME & 4.1 Mini & 58/32 & 0.35 & 16,000 \\
PUPA & 4o-mini & 30/30 & 0.50 & 2,048 \\
PUPA & 4.1 Mini & 30/30 & 0.50 & 2,048 \\
IFBench & 4o-mini & 26/14 & 0.35 & 2,048 \\
IFBench & 4.1 Mini & 26/14 & 0.35 & 4,096 \\
Hearsay & both & 25/25 & 0.50 & 2,048 \\
MMLU (per subject) & 4o-mini & 25/25 & 0.50 & 2,048 \\
MMLU (per subject) & 4.1 Mini & 25/25 & 0.50 & 2,048 \\
\bottomrule
\end{tabular}
\caption{Recorded settings for the matched runs. Solver limit is the configured maximum output tokens, not observed
consumption. Assigned pools precede provider-block exclusions: GPT-4o-mini IFBench evaluates 19 validation items and
GPT-4.1 Mini evaluates 25 mining items. Actual resource use is in Table~\ref{tab:resources}.}
\label{tab:config}
\end{table}

\paragraph{Candidate handling.}
Every changed candidate is committed and becomes the next parent. Best-validation selection compares committed
candidates, breaking ties by mining accuracy and then retaining the earlier candidate; the seed is excluded once a
candidate commits. Eight matched runs used final-round selection instead (GPT-4o-mini PUPA, IFBench, and six MMLU
subjects). Optimization is skipped at seed mining accuracy $\geq 0.95$, except on PUPA. This retained the seed for
GPT-4.1 Mini college mathematics and biology. IFBench with GPT-4.1 Mini proposed no change in round 2.

\FloatBarrier
\subsection{Scoring and Evaluable Items}
\label{app:eval}
\paragraph{Answer extraction.}
MMLU and Hearsay outputs are scored by extracting \texttt{Answer: <letter>} and \texttt{Answer: Yes/No}, respectively,
and comparing with the gold label. AIME uses exact integer match on \texttt{\#\#\#\# <number>} (0--999). Visible
reasoning before these answer markers is not itself scored. IFBench checks only the content following
\texttt{FINAL RESPONSE:}; an item passes when every supported constraint is satisfied. The verifier implements 32
of the 83 constraint types in IFEval and IFBench combined, and evaluated items use the supported types.

\paragraph{PUPA pipeline and metrics.}
The optimized solver prompt rewrites a private query into a redacted request. Fixed GPT-4.1 answers that request;
the solver then synthesizes a final answer under a fixed instruction, using the original query and external response.
Leakage is the fraction of annotated PII strings present in the redacted request. A GPT-5 judge scores final-answer
quality against the reference on a 0--1 scale, with up to 4,096 completion tokens. An item is correct when the
composite $(\text{quality}+1-\text{leakage})/2$ reaches 0.7; reported PUPA accuracy uses this criterion. Only the
rewrite prompt is optimized.

\paragraph{Denominators and accounting.}
Provider-blocked items are excluded rather than scored as incorrect: PUPA evaluates 38 of 40 test items and MMLU
professional law 63 of 66, identically for both solvers. Transitions are computed over common evaluated IDs, so
$\Delta\mathrm{acc}=(|R|-|G|)/|D|$ and $C=|R|+|G|=|N|+2\min(|R|,|G|)$; MMLU transitions sum subjects while its
accuracy is macro-averaged.

\subsection{Related-Work Protocol Comparison}
\label{app:prior}
Table~\ref{tab:prior} expands the protocol comparison in Related Work. Rows differ in task variant, baseline, data,
models, and feedback, so the reported changes describe each protocol rather than rank optimizers.

\begin{table}[htbp]
\centering
\footnotesize
\setlength{\tabcolsep}{3pt}
\begin{tabular}{@{}p{0.18\linewidth}p{0.24\linewidth}p{0.18\linewidth}p{0.18\linewidth}p{0.15\linewidth}@{}}
\toprule
\rowcolor{midblue}Work & Items: optimization / validation / test & Solver models & Feedback and optimizer & Reported change (\%) \\
\midrule
\rowcolor{midgreen}\multicolumn{5}{l}{\textbf{LegalBench-Hearsay}} \\
Trace2Policy, human-written refinements~\citep{zha2026trace2policy} & 30 iteration / 64 held-out, from 94 items & Claude Opus 4.6, Haiku 4.5, Kimi K2.5 & human curator; round 1 also read held-out errors & Haiku 4.5: $79.7\rightarrow 93.8$ \\
Trace2Policy, LLM-written refinements~\citep{zha2026trace2policy} & 61 test items & DeepSeek-v3.2, Kimi K2.5, Kimi K2.6 (mean) & LLM, with a regression gate & $69.4\rightarrow 72.7$ \\
Chen et al.~\citeyearpar{chen2025knowing} (definition classification, a different LegalBench task) & 100 train / 100 test (50 / 50 if under 200 items) & LLaMA-3-8B, LLaMA-3.1-8B, GPT-4o & evaluator-guided rewriting, up to 3 iterations & $55\rightarrow 70$; $56\rightarrow 69$; $83\rightarrow 90$ \\
\rowcolor{softblue}This paper & 25 / 25 / 49, recut from 99 items & GPT-4o-mini, GPT-4.1 Mini & GPT-5, item-level effect report & $71.4\rightarrow 69.4$; $71.4\rightarrow 73.5$ \\
\midrule
\rowcolor{midgreen}\multicolumn{5}{l}{\textbf{MMLU}} \\
MPO~\citep{sharma2026mpo} & 3,000 train / 1,400 test, all subjects & LLaMA-3-8B-Instruct, Mistral-7B-Instruct & critic LLM, section-local gradients & $57.21\rightarrow 61.50$; $53.79\rightarrow 55.50$ \\
\rowcolor{softblue}This paper & six subjects, 25 / 25 / 66 each & GPT-4o-mini, GPT-4.1 Mini & GPT-5, item-level effect report & $75.1\rightarrow 77.2$; $81.0\rightarrow 83.3$ \\
\midrule
\rowcolor{midgreen}\multicolumn{5}{l}{\textbf{AIME}} \\
GEPA~\citep{agrawal2026gepa} & 2022--24 problems (90) split into train and validation; test: 2025, 30 problems, 5 runs each & Qwen3-8B, GPT-4.1 Mini; DSPy chain-of-thought baseline & reflective LLM, Pareto selection & $27.33\rightarrow 32.00$; $49.33\rightarrow 59.33$ \\
\rowcolor{softblue}This paper & 58 / 32 / 30; bare seed prompt & GPT-4.1 Mini & GPT-5, item-level effect report & $53.3\rightarrow 50.0$ \\
\midrule
\rowcolor{midgreen}\multicolumn{5}{l}{\textbf{PUPA}} \\
GEPA~\citep{agrawal2026gepa} & 111 / 111 / 221 & Qwen3-8B, GPT-4.1 Mini & reflective LLM, Pareto selection & $80.82\rightarrow 91.85$; $78.57\rightarrow 94.47$ \\
\rowcolor{softblue}This paper & 30 / 30 / 40 (38 scored) & GPT-4o-mini, GPT-4.1 Mini & GPT-5, item-level effect report & $68.5\rightarrow 79.9$; $68.2\rightarrow 80.7$ \\
\bottomrule
\end{tabular}
\caption{Published evaluation protocols on shared benchmarks. Rows differ in data, solver, baseline prompt, feedback
access, and evaluation budget. Changes follow solver-column order; PUPA changes are composite scores, as reported by GEPA.}
\label{tab:prior}
\end{table}

\FloatBarrier
\section{Prompt and Feedback Specification}
\label{app:prompts}
The specification below comprises the optimizer system prompt, task-specific insertions, and per-round feedback.
The quoted template is the best-validation version; runs using final-round selection received an earlier version
stating that the last round is returned.

\subsection{Optimizer System Prompt}
\label{app:sysprompt}
\texttt{\{task\_description\}} and \texttt{\{max\_rounds\}} are filled per benchmark and run. The template's wording is
preserved, with reflowed lines and ASCII arrows typeset as $\rightarrow$.

\begin{promptbox}[title={\textbf{Optimizer system prompt} (best-validation version)},fontupper=\small]{softblue}
You are working with an expert prompt engineer to help a smaller LLM (the ``solver'') solve \texttt{\{task\_description\}} more
reliably.\par\medskip
\textbf{HOW THIS PROCESS WORKS} (read carefully -- this is a measured, multi-round experiment, not a one-shot rewrite):\par
\noindent 1. You get up to \texttt{\{max\_rounds\}} rewrite rounds. After each round, your rewritten prompt is run on a WORKING SET of
examples and on a VALIDATION set -- a second, disjoint batch meant to approximate unseen data.\par
\noindent 2. From round 2 on, you are shown the measured EFFECT of your previous rewrite, IN FULL, on BOTH sets: every working-set item it
RECOVERED (wrong $\rightarrow$ right), every item it REGRESSED (right $\rightarrow$ wrong) together with the solver's new wrong answer,
the previous text of every section you changed, and -- just as fully -- every VALIDATION item it recovered and every validation item it
regressed, with the same detail. Nothing about validation is held back from you.\par
\noindent 3. In those later rounds your job is causal: first diagnose WHICH of your previous edits caused WHICH regressions and
recoveries on EITHER set, then rewrite -- keep the edits that recovered items, revert or repair the edits that regressed items, and
address the remaining failures. Do not churn wording that was not implicated by the evidence.\par
\noindent 4. There is no ``last round always ships'' rule. Every round commits unconditionally to a version history, and at the end,
whichever COMMITTED round scores BEST on validation accuracy is the one that actually ships -- not necessarily the last round, and not
necessarily this one. Validation accuracy is therefore not just a diagnostic number: it is literally the criterion your rewrite is
competing on. Treat every round as a real, standalone candidate for shipping, not a throwaway step toward some final round.\par
\noindent 5. Because you now see every validation item that flips, a rewrite that merely patches the specific items shown -- without
addressing the underlying rule that produced them -- will look good this round and can still fail on the next genuinely new case.
Prefer fixing the general rule that explains the failures over specific patches to the exact items shown.\par\medskip
You will see: the solver's current markdown-formatted prompt; (from round 2) the EFFECT REPORT of your previous rewrite described above;
EVERY question the solver currently gets wrong on the working set, with the solver's answer and the correct answer alongside (not a
sample -- all of them); and EVERY question the current prompt already gets right on the working set (not a sample -- all of them), so you
know exactly what must not break.\par\medskip
\textbf{CRITICAL RULE ABOUT EXAMPLES} -- READ BEFORE WRITING ANYTHING: you are shown every failure, every success, and every validation
item so you can DIAGNOSE what is going wrong, NOT so you can turn them into illustrative examples. When you write an invented example in
the Context or Task Details section: (a) Do NOT reuse the underlying scenario, fact pattern, or entities of ANY failure, gold, or
validation item shown to you in this conversation -- not verbatim, and not paraphrased with different names. Changing ``Linda and her
editors'' to ``Sarah and her colleagues'' is still copying; the solver pattern-matches on the STORY, not the names. (b) Before writing
an invented example, actively check: does this resemble the setup of anything I was just shown (a lab report, a bystander pointing, an
email admitting knowledge, a customer asking for a refund, someone hearing about a planned crime, a doctor's diagnosis, and so on)? If
yes, discard it and invent a genuinely different scenario. (c) Prefer stating the principle abstractly with NO example at all over
reusing a shown scenario's shape. An abstract rule generalizes; a disguised replay of a shown item does not -- it is memorization with a
different name attached, and it is exactly what causes a prompt to score well on the items you were shown and fail on the next genuinely
new one.\par\medskip
\textbf{CRITICAL FACT ABOUT THE SOLVER:} it has NO hidden scratchpad. Its ONLY reasoning space is the text it actually writes out. So
``think step by step'' works only if the solver WRITES the steps in its output, before the final answer line. Instructions like
``reason internally'', ``use a scratchpad but do not show it'', or ``output only the answer'' DESTROY the reasoning entirely -- for this
model they are identical to ``do not reason'', regardless of task type. Where a task does not need reasoning, the correct instruction is
to keep the answer direct -- never phrasings like these.\par\medskip
\textbf{FIRST, identify the task type from the failures you are shown:} (i) Reasoning tasks -- anything needing calculation, derivation,
formal logic, OR applying a rule/definition to a specific scenario (e.g. mathematics, econometrics, and hearsay / most legal analysis):
the solver RELIABLY does better when it WRITES OUT its working step by step before the final answer line. A bare answer starves it.
(ii) Pure factual-recall / lookup tasks -- where the answer is simply known or not, with no intermediate steps (e.g. a memorised
definition or fact): forced step-by-step reasoning can make the solver over-think and second-guess a fact it already recalled, LOWERING
accuracy. Sharpen the definitions and framing instead, and keep the answer direct. When unsure which it is, ALLOW visible reasoning --
it helps more often than it hurts, and the final answer line is extracted either way. Match your rewrite to the task type before doing
anything else.\par\medskip
\textbf{A good rewrite:} surfaces the general reasoning principle that separates correct from incorrect answers on this task, and states
it in terms the solver can apply to any new case it has not seen before; gives the solver the mental scaffolding it needs -- a clearer
way to think about the problem, a checklist, a decision procedure, a definition it can lean on; illustrates principles with GENUINELY
INVENTED examples when helpful -- different scenario, different entities, different setup from anything you were shown -- or no example
at all; keeps the fixed markdown schema (\texttt{\#\# System Role}, Context, Task Details, Constraints, Output Format), where you may
edit any section but must not add or remove these headers; keeps the FINAL answer line's format exactly (e.g. a line
\texttt{Answer: <LETTER>} or \texttt{Answer: Yes/No}), which is ALL the scorer reads and is searched for anywhere in the output, so you
MAY, and for reasoning tasks SHOULD, let the solver write visible step-by-step working BEFORE it (NEVER add ``output only the answer'',
``do not show your working'', or ``reason internally''); and MATCHES the reasoning style to the task type, with the objective being
accuracy, never verbosity for its own sake.\par\medskip
\textbf{RESPONSE FORMAT:} you MAY begin your reply with a short \texttt{\#\# Analysis} section (2-6 sentences diagnosing which previous
edit caused which regression or recovery -- state the cause before the fix; in round 1 you may use it to state your hypothesis instead).
The \texttt{\#\# Analysis} section is discarded automatically by the parser. After it, output the full rewritten markdown with the five
standard section headers. No other prose, no code fences, nothing after the last section.
\end{promptbox}

\paragraph{No-report variant.}
In the ablation of Appendix~\ref{app:ablation}, the no-report arm receives the same prompt without points 2, 3, and 5,
without the effect-report entry in the list of inputs, and without the references to validation items in the example
rule. Point 4 keeps its shipping rule without the remark on validation accuracy, and the response-format note asks for
a hypothesis about the failures instead of a diagnosis of the previous edit. The user message contains no effect
report and no promise of one. Everything else is identical.

\subsection{Task-Specific Insertions}
\label{app:insertions}
\texttt{\{task\_description\}} supplies fixed task-specific guidance, quoted below. The MMLU
guidance distinguishes computational from recall subjects.

\begin{promptbox}[title={\textbf{Task descriptions inserted into the system prompt}},fontupper=\small]{softgreen}
\textbf{Hearsay.} a binary hearsay classification task under U.S. Federal Rule of Evidence 801 (deciding whether a courtroom statement is
hearsay). The output is Yes or No.\par
\textbf{MMLU.} a 4-way multiple-choice exam question (options A, B, C, D) from one academic subject. Computational subjects (mathematics,
econometrics) need multi-step working; factual-recall subjects (law, computer security, and similar) are usually answered best directly,
without forced step-by-step reasoning. The final answer is a single letter A, B, C, or D on an ``Answer:'' line.\par
\textbf{AIME.} a competition-level (AIME) math problem, substantially harder than grade-school word problems -- multi-step algebra,
combinatorics, number theory, geometry, or probability. The correct answer is always a specific integer between 0 and 999. The output
must end with a line containing the final numeric answer.\par
\textbf{IFBench.} an instruction-following task with the same mechanically verifiable constraint structure and the same
\texttt{FINAL RESPONSE:} marker convention as IFEval (see that description for the full explanation), but drawn from a broader set of constraint types (word/consonant/vowel counts,
keyword-frequency and forbidden-word rules, formatting requirements, and more). Correctness means the text after the marker satisfies
every listed constraint; there is no single gold answer to match.\par
\textbf{PUPA.} a privacy-conscious query-rewriting task: the input is a PRIVATE user request (may contain names, contact info, employers, or
other personal details). You are ONLY editing the REWRITE instruction -- the prompt that tells a model how to anonymize/generalize the
request before it is sent to a separate, untrusted external model. You are NOT editing how the final answer to the user is composed; that
step is fixed elsewhere. A failure means the rewritten request still leaked one or more specific private details (shown to you as
``leaked $N/M$ PII units'') -- the fix is to make the rewrite instruction more systematic about generalizing/abstracting every private
detail (names $\rightarrow$ roles, specific employers $\rightarrow$ generic descriptions, exact locations $\rightarrow$ broader regions,
etc.) while still preserving enough of the request's substance that the external model's answer stays useful. The ``gold''/quality signal
shown to you reflects the FINAL answer's usefulness (not directly under your control), so do not sacrifice the request's meaning purely to
hide details -- an unusable rewrite scores just as badly as a leaky one.
\end{promptbox}

\subsection{Seed Prompt Example}
\label{app:seed}
The AIME seed illustrates the five-section schema. Other benchmark seeds are included in the source archive.
Only line breaks are adjusted below.
{\small
\begin{verbatim}
## System Role
You are a competition mathematician.

## Context
You will be given a competition-style math problem whose answer is always
an integer between 0 and 999.

## Task Details
Solve the problem.

## Constraints
The final answer must be an integer between 0 and 999.

## Output Format
After your work, write the final numeric answer on its own line in exactly
this form:
#### <number>
\end{verbatim}
}

\subsection{Per-Round Feedback}
\label{app:feedback}
The user message contains the iteration number and budget, the current prompt, every mining failure (question, wrong
output, gold answer), and every mining success (question and gold). From round 2, its effect report adds the previous
text of changed sections, recovered and regressed items on both $M$ and $V$, and before/after validation accuracy.
Recovered items include question and gold; regressed items also include the new wrong output, truncated to its last
400 characters. The message ends with a request to diagnose the observations and return the five-section markdown.
The parser discards an optional \texttt{\#\# Analysis} preamble; a no-change response does not commit a new version.
IFBench pins Output Format to preserve its answer marker; the other four sections remain editable.

\FloatBarrier
\section{Detailed Results and Trajectories}
\label{app:results}
The following records expand the aggregate results: paired accuracy and test transitions, then round-level tuning
outcomes, followed by prompt excerpts illustrating the recorded revisions.

\subsection{Paired Accuracy and Test Transitions}
\label{app:gap}
Table~\ref{tab:pergap} lists all 19 matched records, including each MMLU subject. It combines the tuning/test
accuracy pairs with the test recoveries and regressions. Summing the subject counts gives the MMLU totals in
Table~\ref{tab:main}. ``Edited'' distinguishes returned candidates from retained seeds, not an improvement gate.

\begin{table}[htbp]
\centering
\small
\setlength{\tabcolsep}{3pt}
\begin{tabular}{llccrcrc}
\toprule
\rowcolor{midblue}Run & Solver & Edited & \shortstack{Validation (\%)\\seed $\rightarrow$ returned} & $\Delta V$ & \shortstack{Test (\%)\\seed $\rightarrow$ returned} & $\Delta T$ & Rec./Reg. \\
\midrule
AIME & 4.1 Mini & yes & 62.5 $\rightarrow$ 68.8 & $+6.2$ & 53.3 $\rightarrow$ 50.0 & $-3.3$ & 3/4 \\
\rowcolor{softgreen}PUPA & 4o-mini & yes & 40.0 $\rightarrow$ 53.3 & $+13.3$ & 55.3 $\rightarrow$ 65.8 & $+10.5$ & 7/3 \\
\rowcolor{softgreen}PUPA & 4.1 Mini & yes & 46.7 $\rightarrow$ 70.0 & $+23.3$ & 44.7 $\rightarrow$ 71.1 & $+26.3$ & 10/0 \\
\rowcolor{softgreen}IFBench & 4o-mini & yes & 36.8 $\rightarrow$ 52.6 & $+15.8$ & 40.5 $\rightarrow$ 45.2 & $+4.8$ & 6/4 \\
\rowcolor{softgreen}IFBench & 4.1 Mini & yes & 57.1 $\rightarrow$ 71.4 & $+14.3$ & 42.9 $\rightarrow$ 66.7 & $+23.8$ & 12/2 \\
Hearsay & 4o-mini & yes & 68.0 $\rightarrow$ 80.0 & $+12.0$ & 71.4 $\rightarrow$ 69.4 & $-2.0$ & 8/9 \\
\rowcolor{softgreen}Hearsay & 4.1 Mini & yes & 76.0 $\rightarrow$ 96.0 & $+20.0$ & 71.4 $\rightarrow$ 73.5 & $+2.0$ & 7/6 \\
\midrule
\rowcolor{softgreen}MMLU college math. & 4o-mini & yes & 84.0 $\rightarrow$ 80.0 & $-4.0$ & 75.8 $\rightarrow$ 77.3 & $+1.5$ & 6/5 \\
MMLU computer sec. & 4o-mini & yes & 72.0 $\rightarrow$ 80.0 & $+8.0$ & 90.9 $\rightarrow$ 89.4 & $-1.5$ & 1/2 \\
\rowcolor{softgreen}MMLU econometrics & 4o-mini & yes & 60.0 $\rightarrow$ 72.0 & $+12.0$ & 66.7 $\rightarrow$ 72.7 & $+6.1$ & 10/6 \\
\rowcolor{softgreen}MMLU high sch.\ biol. & 4o-mini & yes & 92.0 $\rightarrow$ 96.0 & $+4.0$ & 87.9 $\rightarrow$ 89.4 & $+1.5$ & 2/1 \\
\rowcolor{softgreen}MMLU philosophy & 4o-mini & yes & 80.0 $\rightarrow$ 84.0 & $+4.0$ & 77.3 $\rightarrow$ 80.3 & $+3.0$ & 5/3 \\
\rowcolor{softgreen}MMLU prof.\ law & 4o-mini & yes & 59.1 $\rightarrow$ 68.2 & $+9.1$ & 52.4 $\rightarrow$ 54.0 & $+1.6$ & 7/6 \\
\midrule
MMLU college math. & 4.1 Mini & no & -- & 0.0 & 92.4 $\rightarrow$ 92.4 & 0.0 & 0/0 \\
\rowcolor{softgreen}MMLU computer sec. & 4.1 Mini & yes & 80.0 $\rightarrow$ 76.0 & $-4.0$ & 87.9 $\rightarrow$ 90.9 & $+3.0$ & 3/1 \\
MMLU econometrics & 4.1 Mini & yes & 80.0 $\rightarrow$ 80.0 & 0.0 & 81.8 $\rightarrow$ 80.3 & $-1.5$ & 1/2 \\
MMLU high sch.\ biol. & 4.1 Mini & no & -- & 0.0 & 87.9 $\rightarrow$ 87.9 & 0.0 & 0/0 \\
\rowcolor{softgreen}MMLU philosophy & 4.1 Mini & yes & 88.0 $\rightarrow$ 88.0 & 0.0 & 80.3 $\rightarrow$ 86.4 & $+6.1$ & 4/0 \\
\rowcolor{softgreen}MMLU prof.\ law & 4.1 Mini & yes & 81.8 $\rightarrow$ 81.8 & 0.0 & 55.6 $\rightarrow$ 61.9 & $+6.3$ & 9/5 \\
\bottomrule
\end{tabular}
\caption{All matched accuracy pairs and test transition counts. Deltas are percentage points computed before rounding;
green rows have positive test deltas. MMLU subjects have 66 assigned test items (63 evaluated for professional law).
Dashes denote unmeasured validation after a high-baseline skip; zero $\Delta V$ denotes the retained seed.}
\label{tab:pergap}
\end{table}

\FloatBarrier
\subsection{Round-by-Round Trajectories}
\label{app:rounds}
Table~\ref{tab:rounds} records the seven matched non-MMLU trajectories. Round 0 is the seed; each later row compares
with the preceding committed prompt, not with the seed. Stars mark best-validation returns and daggers final-round
returns. MMLU subject-level endpoints remain in Table~\ref{tab:pergap}; full committed trajectories are in the analysis records.

\begin{table}[htbp]
\centering
\small
\setlength{\tabcolsep}{4pt}
\begin{tabular}{llrccccc}
\toprule
\rowcolor{midblue}Benchmark & Solver & Round & $M$ acc. & $V$ acc. & $M$ rec./reg. & $V$ rec./reg. & Sections changed \\
\midrule
AIME & 4.1 Mini & 0 & 46.6 & 62.5 & -- & -- & -- \\
 & & 1$^*$ & 51.7 & 68.8 & 9/6 & 5/3 & Con, TD \\
 & & 2 & 48.3 & 65.6 & 6/8 & 2/3 & TD \\
 & & 3 & 46.6 & 59.4 & 6/7 & 3/5 & TD \\
\midrule
PUPA & 4o-mini & 0 & 69.0 & 40.0 & -- & -- & -- \\
 & & 1 & 79.3 & 56.7 & 6/3 & 5/0 & Con, Ctx, TD \\
 & & 2 & 86.2 & 63.3 & 4/2 & 5/3 & TD \\
 & & 3$^\dagger$ & 82.8 & 53.3 & 1/2 & 0/3 & TD \\
\midrule
PUPA & 4.1 Mini & 0 & 72.4 & 46.7 & -- & -- & -- \\
 & & 1 & 75.9 & 63.3 & 4/3 & 6/1 & Con, Ctx, TD \\
 & & 2 & 69.0 & 63.3 & 1/3 & 0/0 & TD \\
 & & 3$^*$ & 75.9 & 70.0 & 3/1 & 3/1 & TD \\
\midrule
IFBench & 4o-mini & 0 & 40.0 & 36.8 & -- & -- & -- \\
 & & 1 & 60.0 & 52.6 & 4/0 & 4/1 & Con, Ctx, OF, SR, TD \\
 & & 2 & 55.0 & 68.4 & 0/1 & 3/0 & TD \\
 & & 3$^\dagger$ & 50.0 & 52.6 & 0/1 & 0/3 & TD \\
\midrule
IFBench & 4.1 Mini & 0 & 48.0 & 57.1 & -- & -- & -- \\
 & & 1$^*$ & 52.0 & 71.4 & 3/2 & 2/0 & Con, Ctx, SR, TD \\
 & & 2 & \multicolumn{5}{c}{optimizer proposed no change} \\
 & & 3 & 40.0 & 71.4 & 2/5 & 0/0 & Con \\
\midrule
Hearsay & 4o-mini & 0 & 72.0 & 68.0 & -- & -- & -- \\
 & & 1 & 68.0 & 44.0 & 3/4 & 1/7 & Con, Ctx, TD \\
 & & 2 & 68.0 & 72.0 & 2/2 & 7/0 & Ctx, TD \\
 & & 3$^*$ & 76.0 & 80.0 & 3/1 & 2/0 & TD \\
\midrule
Hearsay & 4.1 Mini & 0 & 80.0 & 76.0 & -- & -- & -- \\
 & & 1 & 88.0 & 92.0 & 2/0 & 4/0 & Con, Ctx, TD \\
 & & 2$^*$ & 80.0 & 96.0 & 1/3 & 1/0 & Ctx, TD \\
 & & 3 & 80.0 & 84.0 & 1/1 & 0/3 & TD \\
\bottomrule
\end{tabular}
\caption{Round-by-round trajectories (accuracy in \%). Sections: SR system role, Ctx context, TD task details,
Con constraints, OF output format. In IFBench with GPT-4.1 Mini, rounds 1 and 3 tie on validation and round 1
is returned on its higher mining accuracy.}
\label{tab:rounds}
\end{table}

Across the 20 committed rounds in Table~\ref{tab:rounds}, nine regressed more mining items than they recovered and six
lowered validation accuracy.

\FloatBarrier
\subsection{Worked Prompt Changes}
\label{app:evolution}
The excerpts below show the returned Task Details of two runs. Context and Constraints also changed in Hearsay;
System Role and Output Format were unchanged there, and IFBench's Output Format was pinned.

\paragraph{Hearsay, GPT-4.1 Mini.}
The seed's Task Details consisted of ``Decide whether the given statement is hearsay.'' The returned round-2 section
is reproduced below with formatting adjusted. Validation changed from 76.0\% to 96.0\%, while test accuracy changed
from 35/49 to 36/49: seven recoveries and six regressions. The explicit procedure raised validation sharply while
moving thirteen test outcomes for a one-item net gain.

\begin{promptbox}[title={\textbf{Hearsay: returned Task Details} (round 2 excerpt)},fontupper=\small]{softgreen}
Decide whether the given statement is hearsay by applying this decision procedure and explain your steps
briefly before the final answer:\par
\textit{Step 1.} Is there a ``statement''? Identify whether the evidence is an assertion (oral/written) or assertive conduct.
Non-assertive words (questions, commands, offers) and non-assertive conduct are not statements.\par
\textit{Step 2.} Was it made out of court? In-court testimony in this proceeding is not hearsay for this purpose.\par
\textit{Step 3.} What specific proposition is the evidence offered to prove (ultimate or intermediate)? Then apply the independent
relevance test: would the evidence still be probative of that proposition even if the content of the statement were false? If yes, it
is offered for a non-truth purpose; if no, it is offered for its truth.\par
\textit{Step 4.} If the evidence is offered solely for a non-truth purpose, answer No. Common non-truth purposes: effect on the listener,
notice, knowledge or motive where only the hearing of the words matters; proving that a conversation or words occurred; and
circumstantial use of words to show a condition or state of the speaker unrelated to their truth.\par
\textit{Step 5 (guardrail on embedded assertions).} If the proponent relies on the statement to establish an intermediate fact that
itself depends on the truth of an asserted proposition within it (e.g., ``I heard/know/saw X''), the statement is used for its
truth; treat as hearsay. Do not invoke the guardrail when the use is only to show that words were spoken, the speaker's state, or an
effect on the listener where content truth is irrelevant under Step 3.\par
\textit{Step 6.} If out-of-court and offered for the truth of what it asserts, answer Yes; otherwise answer No.
\end{promptbox}

\paragraph{IFBench, GPT-4.1 Mini.}
The seed's Task Details was ``Answer the user's request.'' Round 1 rewrote four sections, adding constraint extraction,
planning, and verification; the excerpt below shows its Task Details. Validation changed from 57.1\% to 71.4\%, and
test accuracy from 18/42 to 28/42 (twelve recoveries, two regressions). The \texttt{FINAL RESPONSE:} marker stayed fixed,
so only the post-marker answer was checked.

\begin{promptbox}[title={\textbf{IFBench: returned Task Details} (round 1 excerpt)},fontupper=\small]{softblue}
Your job is to: read the prompt and explicitly extract every constraint; plan content that satisfies all constraints
exactly; verify counts and formats on your draft before producing the final response. Prefer visible reasoning for these rule-application
tasks: write your working steps before the final answer line so you can check and fix issues. Only the content after the final marker is
scored.
\end{promptbox}

\FloatBarrier
\subsection{Effect-Report Ablation}
\label{app:ablation}
Both arms use the seed prompt, splits (25 mining, 25 validation, and 49 test items), solver, optimizer, and selection
rule of the main Hearsay GPT-4o-mini run. Each run replays that run's round-1 prompt, which changes Context, Task
Details, and Constraints, and then runs rounds 2 and 3. The arms differ only in what the optimizer is shown: with the
report it receives the item-level effect of its previous edit; without it, it receives the current failures and
successes only, and the report-specific instructions are removed from its system prompt (Appendix~\ref{app:sysprompt}).
We ran five runs per arm and alternated which arm ran first. Table~\ref{tab:ablation_rounds} gives edit sizes and item
transitions per round; Table~\ref{tab:ablation_runs} gives run-level means.

\begin{table}[htbp]
\centering
\small
\setlength{\tabcolsep}{3.5pt}
\begin{tabular}{llcccc}
\toprule
\rowcolor{midblue}Round & Arm & Sections changed & Single-section & \shortstack{Mining\\rec./reg.} & \shortstack{Validation\\rec./reg.} \\
\midrule
2 & report & 2.00 & 1/5 & 12/7 & 26/2 \\
 & no report & 1.60 & 3/5 & 17/14 & 26/6 \\
\midrule
3 & report & 1.00 & 5/5 & 5/7 & 10/3 \\
 & no report & 2.00 & 0/5 & 10/11 & 18/12 \\
\midrule
2--3 & report & 1.50 & 6/10 & 17/14 & 36/5 \\
 & no report & 1.80 & 3/10 & 27/25 & 44/18 \\
\bottomrule
\end{tabular}
\caption{Edit sizes and item transitions of the five runs per arm, summed over runs. Round 3 starts from different
round-2 prompts in the two arms, so only round 2 compares edits from an identical parent prompt.}
\label{tab:ablation_rounds}
\end{table}

\begin{table}[htbp]
\centering
\small
\setlength{\tabcolsep}{4pt}
\begin{tabular}{lrrr}
\toprule
\rowcolor{midblue}Run-level mean & Report & No report & $p$ \\
\midrule
Mining items changing outcome, per round & 3.1 & 5.2 & 0.024 \\
Mining net gain, per round & 0.3 & 0.2 & 1.000 \\
Mining regressions, per round & 1.4 & 2.5 & 0.024 \\
Validation items changing outcome, per round & 4.1 & 6.2 & 0.206 \\
Validation gain (points) & 4.8 & 2.4 & 0.571 \\
Test items changing outcome, per run & 14.2 & 17.2 & 0.270 \\
Test regressions, per run & 8.4 & 10.0 & 0.127 \\
Test change from seed (points) & $-5.3$ & $-5.7$ & 1.000 \\
\bottomrule
\end{tabular}
\caption{Run-level means over five runs per arm. Per-round metrics average the two optimizer rounds of a run. $p$ is an
exact two-sided permutation test on run means, uncorrected for multiple comparisons.}
\label{tab:ablation_runs}
\end{table}

Validation and test differences between the arms are within run-to-run variation.

\FloatBarrier
\subsection{Additional AIME Runs}
\label{app:aime_extra}
Two earlier AIME runs used a 2,048-token output limit and the earlier final-round selection rule, and are excluded
from the matched analyses. With Claude Haiku~4.5~\citep{anthropic2025haiku45} as solver, three rounds were committed and
the third was returned; validation accuracy rose from 62.5\% to 78.1\% and test accuracy from 26.7\% to 33.3\% (8/30 to
10/30, two recoveries and no regressions). With GPT-4o-mini, the last round's validation accuracy (12.5\%) fell below the
baseline's (15.6\%), so the seed prompt was returned. Its two test evaluations of this identical prompt solved 4 and 3 of
30 problems. Table~\ref{tab:aime_extra} also lists later re-evaluations of that prompt, and of the round-1 prompt returned
in the GPT-4.1 Mini run (Table~\ref{tab:main}), with GPT-4o-mini as solver. The 16,000-token limit used for GPT-4.1 Mini
solved one more problem than the 2,048-token limit, and the GPT-4.1 Mini prompt regressed three problems against the
baseline at the same limit.

\begin{table}[htbp]
\centering
\small
\setlength{\tabcolsep}{4pt}
\begin{tabular}{llrr}
\toprule
\rowcolor{midblue}Solver & Evaluated prompt & Token limit & Test correct \\
\midrule
Claude Haiku 4.5 & seed & 2,048 & 8/30 \\
 & returned (round 3) & 2,048 & 10/30 \\
\midrule
GPT-4o-mini & seed & 2,048 & 4/30 \\
 & returned (the seed, unchanged) & 2,048 & 3/30 \\
 & seed, re-evaluated & 2,048 & 4/30 \\
 & seed, re-evaluated & 16,000 & 5/30 \\
 & GPT-4.1 Mini round-1 prompt & 16,000 & 2/30 \\
\bottomrule
\end{tabular}
\caption{Problems solved out of the 30 AIME 2025 test problems in the earlier runs and in re-evaluations of fixed
prompts. The GPT-4o-mini returned prompt is identical to its seed, so the 4/30 and 3/30 rows differ only by
re-evaluation noise.}
\label{tab:aime_extra}
\end{table}

\FloatBarrier
\section{Evaluation Checks and Artifacts}
\label{app:noisebugs}

\subsection{Re-evaluation of an Unchanged Prompt}
\label{app:noise}
A GPT-4o-mini IFBench run that returned no edited prompt re-evaluated an unchanged prompt on the same 42 test items:
one outcome changed (no recoveries, one regression). The edited prompt of the main GPT-4o-mini IFBench run, which
started from a different seed prompt, changed ten outcomes (six recoveries, four regressions). A GPT-4o-mini AIME run
that returned its seed prompt shows the same effect: two evaluations of the identical prompt solved 4 and 3 of 30
problems (Appendix~\ref{app:aime_extra}).

\subsection{Test-Exposure Diagnostic}
\label{app:oracle}
A diagnostic on LegalBench-Hearsay used an earlier variant of the optimizer with GPT-5, seed 42, two rounds, 30
training items, and 64 test items. With the test items held out, accuracy changed from 44/64 to 47/64 (68.8\% to
73.4\%). Supplying the same test items as optimizer feedback raised accuracy to 61/64 (95.3\%), a gain more than five
times the held-out gain. This run is excluded from all other analyses.

\subsection{Verbatim-Overlap Check}
\label{app:overlap}
For each run we lowercase the returned prompt (the seed when none was returned), keep alphanumeric tokens, and intersect
its word 8-grams with all string fields in the train and test dataset files, including answer-option text. Sixteen of
the 19 prompts share none, and three share some (Table~\ref{tab:overlap}). The mathematical phrase occurs only in the
test file; the True/False template occurs in both train and test.

\begin{table}[htbp]
\centering
\small
\setlength{\tabcolsep}{3pt}
\begin{tabular}{llrrl}
\toprule
\rowcolor{midblue}Run & Solver & Prompt & Shared & Example (lowercased) \\
\midrule
MMLU college math. & 4o-mini & 885 & 1 & the set of all finite subsets of r \\
MMLU computer sec. & 4o-mini & 655 & 5 & false false c true false d false true \\
MMLU computer sec. & 4.1 Mini & 1033 & 5 & false false c true false d false true \\
\bottomrule
\end{tabular}
\caption{Runs with any shared 8-gram. ``Prompt'' is the number of distinct 8-grams in the returned prompt. The shared
strings are a generic mathematical phrase and a True/False answer-option pattern.}
\label{tab:overlap}
\end{table}
The optimizer prompt also prohibits reusing the scenario of any shown item (Appendix~\ref{app:prompts}).

\FloatBarrier
\subsection{Artifacts and Reproducibility}
\label{app:artifacts}
The anonymous source archive contains the implementation, seed prompts, the analysis scripts that produce every table,
tests, the dependency lock, and setup instructions; credentials, endpoints, local paths, repository history, and
dataset corpora are excluded. Per-item traces are retained by the authors; any release will exclude personal
information, predictions, and gold text.

\end{document}